*Research Article*

# An Ontology-Based Artificial Intelligence Model for Medicine Side-Effect Prediction: Taking Traditional Chinese Medicine as an Example

**Yuanzhe Yao,[1] Zeheng Wang [1,2] Liang Li,[1] Kun Lu,[3] Runyu Liu,[1] Zhiyuan Liu,[1] and Jing Yan[4]**

[1]School of Information and Software Engineering, University of Electronic Science and Technology of China, Chengdu 610054, China
[2]School of Electrical Engineering and Telecommunications, University of New South Wales, Sydney, NSW 2052, Australia
[3]Faculty of Medicine, Ludwig Maximilian University of Munich, Munich 81377, Germany
[4]The First Clinical Medical College, Zhejiang Chinese Medicine University, Hangzhou 310053, China

Correspondence should be addressed to Zeheng Wang; zenwang@outlook.com

In this work, an ontology-based model for AI-assisted medicine side-effect (SE) prediction is developed, where three main components, including the drug model, the treatment model, and the AI-assisted prediction model, of the proposed model are presented. To validate the proposed model, an ANN structure is established and trained by two hundred forty-two TCM prescriptions. These data are gathered and classified from the most famous ancient TCM book, and more than one thousand SE reports, in which two ontology-based attributions, hot and cold, are introduced to evaluate whether the prescription will cause SE or not. The results preliminarily reveal that it is a relationship between the ontology-based attributions and the corresponding predicted indicator that can be learnt by AI for predicting the SE, which suggests the proposed model has a potential in AI-assisted SE prediction. However, it should be noted that the proposed model highly depends on the sufficient clinic data, and hereby, much deeper exploration is important for enhancing the accuracy of the prediction.

## 1. Introduction

Artificial intelligence is a modern technology that is utilized in various fields of medicine [1–3]. At the meantime, traditional Chinese medicine (TCM) is now widely considered as a promising alternative medicine for complementary treatment in cancers or chronic diseases due to the effective methodology practically developed by generations of doctors for almost 4000 years [4]. Based on previous verification, it is undeniable that there are many correlations between the TCM syndromes and western diseases, turning out novel approaches for enhancing the treatment efficiency and developing medicines regarding TCM methodologies [5]. Unfortunately, hindered by the remarkable gap between the modern informatics and the fundament of TCM—ancient

Chinese philosophy, such correlations are still too elusive to be formulated precisely.

Recently, in order to figure out the deep connection between modern science and TCM, the research combining TCM with AI for valid knowledge acquisition and mining attracts great attention, thereby leading to many profound works, such as ontology information system design [6], latent tree models design [7], TCM warehouse for AI application [8], and digital knowledge graph development [2]. Especially, in the view of algorithms, these AI-assisted techniques can be recognized by two different approaches: pattern classification and knowledge mining. The former technology attempts to recognize the correct pathological information such as pulse condition [9–14] and tongue diagnosis [15] of an individual patient. However, the later one, knowledge mining, mainly focuses on finding out various kinds of hidden relationships in the knowledge, for example, the relationships between symptom and symptom, symptom and syndrome, and syndrome and disease [16–20]. In addition, it should be noted that there are many other studies that deserve attention as well, such as classifying herbs by the convolutional neural network model [21] using the deep learning mode to explore the relationship between herbal property and action [22].

On the contrary, researchers face, however, many difficulties in setting up AI for TCM in terms of directly interpreting the TCM semantic system (almost recorded by ancient Chinese doctrines) into structured database. However, in this way, considerable workload must be undertaken by limited numbers of experts who are proficient in both AI and TCM to translate the TCM terminologies and then formulate the modern model thereof. In contrast, as shown in Figure 1, using TCM methodology, but not the modern one, in overcoming the barriers of modern science, designing new medicine, for example, is relatively lacking and thus of significant worth to explore.

In this paper, an ontology-based model is developed to train AI for drug side-effect (SE) prediction, in which the methodology of TCM including syndromes differentiation is applied to determine the ontology-based attributions and optimize the AI components, and consequently, form a novel scheme of effectively predicting the medicine's attribution. Here, limited by the shortage of accurate clinic experiment data of modern medicine, TCM data in famous ancient books are used to verify the model which shows a tremendous potential in medicine discovery. The paper is organized as follows: in Section 2, three main components, including the drug model, the treatment model, and the AI-assisted prediction model, are established to introduce how to use TCM theory to explore the modern drugs; then in Section 3, an artificial neural network- (ANN-) based AI model is established and trained by the collected data; and in Section 4, the prediction performance in the proposed model framework is shown and discussed.

## 2. Methodology

*2.1. Ontology-Based Drug Model.* The artificial intelligence model proposed in this paper is based on ontology that considers the essence of a certain entity as a combination of several fundamental attributions with corresponding values and relationships [2, 23, 24]. Such attributions are not only the definite properties which are already completely recognized by researchers but also the latent properties including unknown information and relationships.

For example, as shown in Figure 2, each drug has certain attributions including the definite ones and latent others, which are all involved in a certain prescription with sufficient records of clinical effects. In addition, assuming our prepared ontology system is complete and exclusive, a new drug or prescription which contains attributions we have already recorded can be depicted easily in the ontology-based semantic system, where we could focus on the superficial relationship between such attributions, or in another word labels, and effects caused thereby. In other words, we avoid literally figuring out the ingredient and other deeper properties of each attribution in the new drug, and therefore, the attribution-effect pair is crucial and could be easily converted into an AI scheme such as ANN to handle the prediction of the treatment. Moreover, the proposed ontology-based attribution model could be revised by more accurate clinic records automatically with AI assistance due to the intentionally fuzzy and dynamic defined latent attributions. In this paper, as discussed later, two items, hot and cold, are presented as the fundamental attributions of any medicines.

*2.2. Ontology-Based Treatment Model.* Based on the proposed drug model, it can be depicted, as in Figure 3, that the model of the treatment procedure via a certain prescription $X$. This prescription contains several drugs including the attributions of known ingredients and the latent attributions. As shown in Figure 3, the latent attributions own the capability of influencing the group of indicators with different unknown paths and efficiencies. In another word, in this model, the results of the treatment of a certain patient X that is defined as the positive or negative change of the corresponding indicator are the comprehensive synthesis of the effects induced by various latent attributions. Therefore, this procedure could be interpreted into TCM-based semantic entities: attribution-indicator pairs performing the effects. It should be noted that the different attributions maybe dominate in influencing the same indicator. Furthermore, the model is compatible with the known ingredients or explored attributions and the effects thereof.

*2.3. AI-Assisted Prediction Model.* Based on the aforementioned drug/prescription and treatment models, as illustrated in Figure 4, the SE prediction of new drug X is realized by comprehensive consideration of the involved ontology-based latent attributions with their influential factors (IFs) revised by sufficient medicines' clinic records that contain, for instance, the attribution no. 3 and X, where the revision procedure could be undertaken by an AI scheme such as ANN. Also, the same AI scheme could predict the SE with the trained pattern.

It should be noted that the IFs must be linked with the corresponding attributions and indicators which means the trained model is consisted of IFs' indicator vectors but not the isolated IFs as the input. In this way, the ontology-based model that the latent attributions with corresponding IFs influence a certain indicator is established. Next, we will generate an AI scheme to validate our proposed model by determining two latent attributions, which are hot and cold of the prescription, and a simple indicator: whether the prescription causes SE or not when this prescription is used in a right way.

## 3. Experiment Detail

According to the analysis in Section 2, it is the key for establishing the proposed model that determines the attributions and obtains the IFs' indicator vectors. However, owing to the lack of related theory, generating the

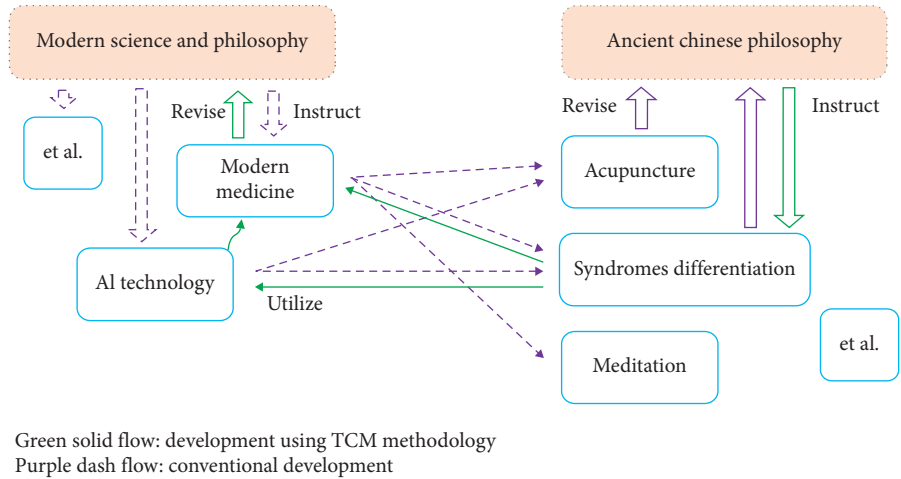

Green solid flow: development using TCM methodology
Purple dash flow: conventional development attributions directly, comprehensively, and exclusively is very hard. Therefore, we follow the theory of TCM which has the advantage in the matured ontology-based semantic system that can determine the attributions spontaneously. For example, hot and cold are two main attributions catalogized by TCM theory, where all the drugs must contain one out of these two attributions, leading to a charming approach for determining the latent attributions of western drugs in the same way.

Figure 1: The development procedure based on modern science and the TCM-based ontology.

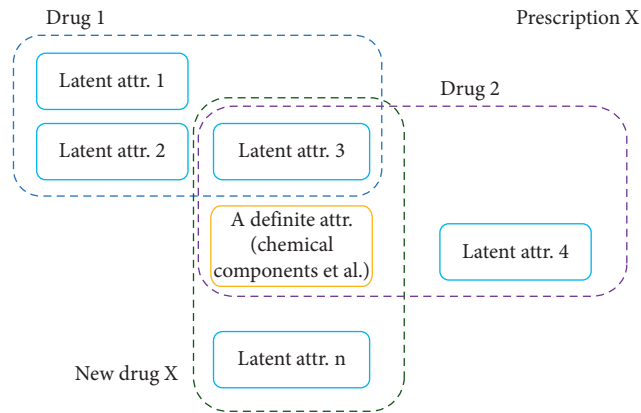

Figure 2: Ontology-based drug model and latent attributions thereof.

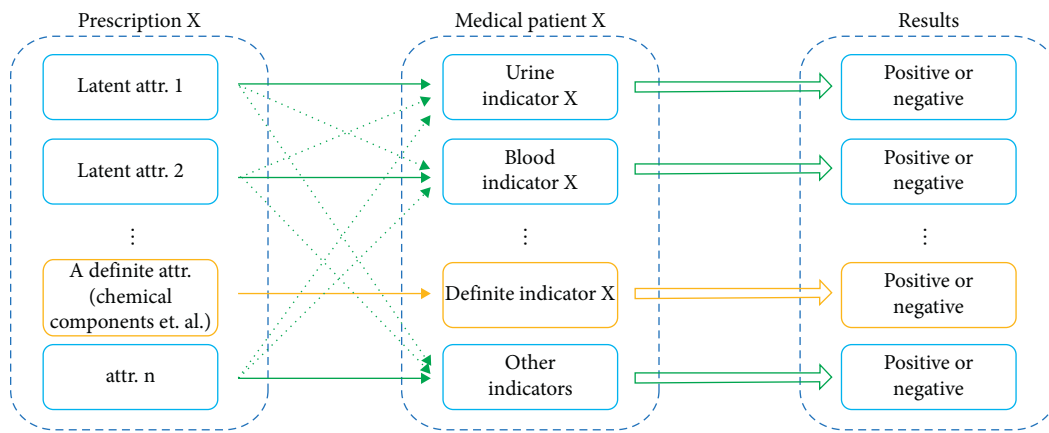

Figure 3: Ontology-based treatment model concerning the attribution-indicator relationships.

As shown in Figure 5, after the identification of the attributions and indicators, we should establish and train the AI model. Here, we gathered the detailed data, including 150 effective prescriptions, the dosages thereof, and the corresponding indicators from the famous ancient TCM book

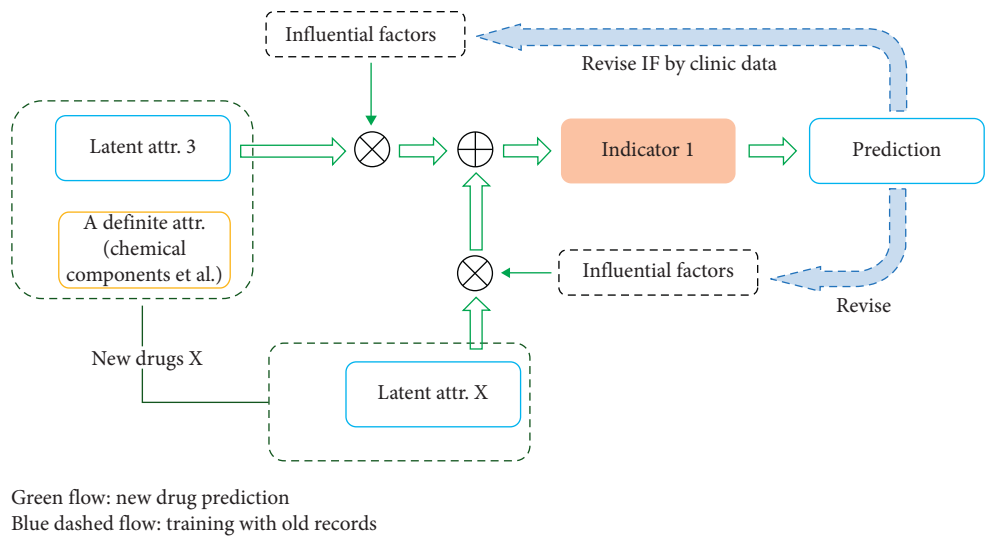

Figure 4: The network for training AI using proposed models.

*Shanghanzabinglun (Treatise on Cold Pathogenic and Miscellaneous Diseases)* which is considered as the origin of practical TCM prescription in clinic. In addition, as concluded before, according to the practice identification by ancient TCM doctors and the TCM standards published by Chinese government [25, 26], we labeled two ontology-based attributions that are hot and cold for describing the drugs' fundamental property, which is the first step of conducting the prediction as depicted in Figure 5. Thereafter, we assigned the IFs of each attribution equaling the total dosage of the drugs which own the corresponding attribution in the prescription. Since in the ontology-based labeling procedure, a drug must belong to one certain catalogue out of the two in total, using the summarized dosage to represent the IFs is reasonable; however, it needs more verification in future research. In addition, it should be noted that some prescriptions do not contain any drugs (for example, some uncatalogued pure chemical ingredients) associated with hot or cold, where for convenience these drugs could be considered as neutral ones and not affiliated with the two attributions mentioned before.

As shown in Figure 6(a), 242 effective prescriptions are dotted regarding the normalized total dosage, where the $x$-axis and $y$-axis in the figure represent the total hot dosage and the total cold dosage, respectively. These dosages are considered as the IF factors of the prescription. According to our best knowledge, because there are no reports indicating the 150 prescriptions gathered from the book *Shanhanzabinglun*, we consider these prescriptions are the safe prescriptions. In contrast, we gathered 92 unsafe prescriptions reported to frequently cause SE when they are used in a right way.

The distribution of the percentages of the safe prescriptions and the unsafe prescriptions features a huge difference in terms of whether the dosage is stronger than 500, which suggests the reported unsafe prescriptions own the characteristics that can be distinguished from the safe prescriptions. Therefore, we try to use the pattern recognition

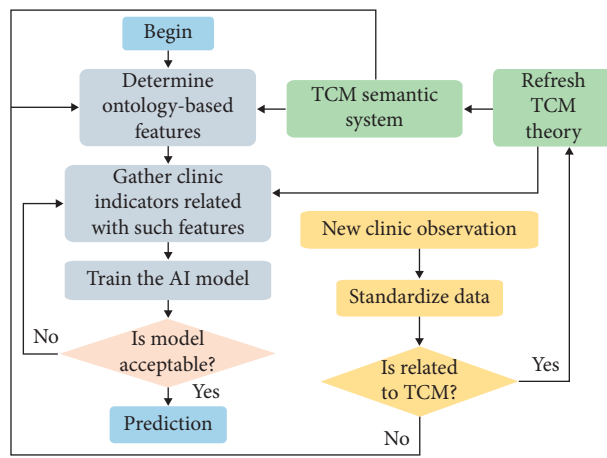

Figure 5: The SE prediction procedure of the proposed model.

method to build a simple classifier to predict which prescription is unsafe.

ANN is a classic model in pattern recognition tasks. Due to its good performance and simple form, it is widely used in solving nonlinear classification problems. Here, a multilayer ANN model is developed to learn how to recognize the special pattern from our collected prescription data.

In order to use the ANN model to train this classifier, we represent each prescription into a vector. We analyze each prescription and identify dosages about every single herbal drug which form the prescription. According to "Chinese Pharmacopoeia," we can clarify hot/cold properties of each drug appearing in our collected prescriptions. We use the weighted BOW model to represent prescription, $v_p$ [27, 28]. Furthermore, we generate a weighted matrix according to the BOW model, $W$. The matrix $W$ has two columns, and each column in $W$ represent a type of property and each row in $W$ represent the hot or cold property on a single drug. We use this model-generated matrix as a linear operator to generate the input vector in

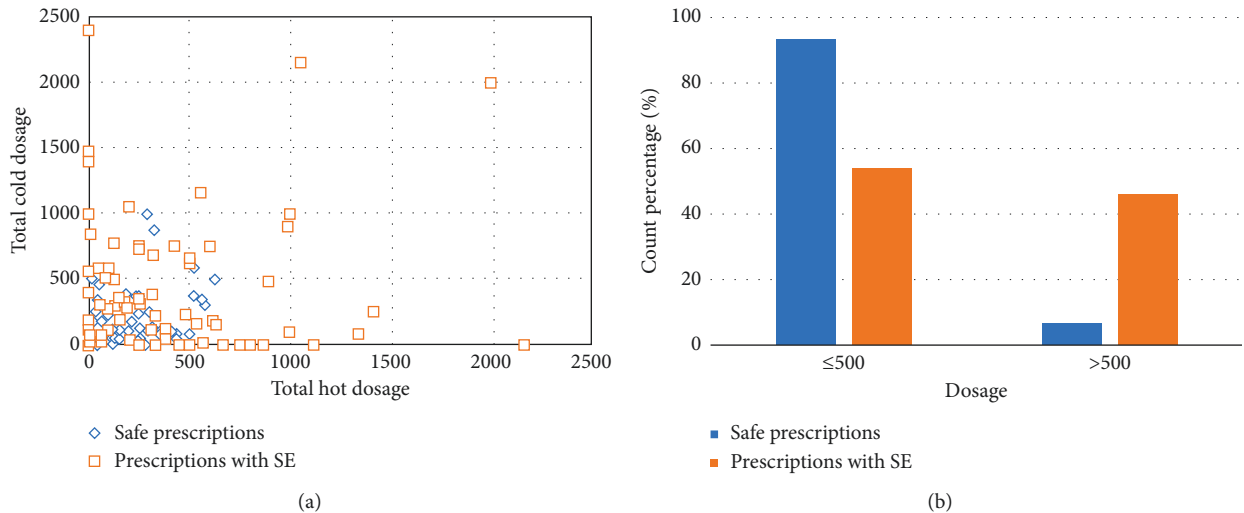

Figure 6: The counts of the hot/cold IF (counts) in the book.

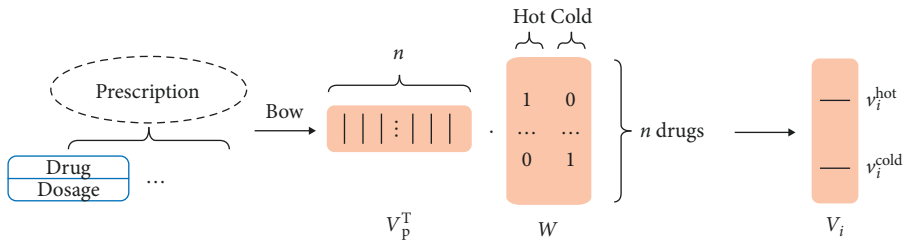

Figure 7: The schematic procedure of converting prescription into a vector.

the ANN model. Thus, the input vector of the ANN model can be expressed as follows:

$$v_i = W^T \cdot v_p, \qquad (1)$$

where $v_i$ is the input vector in the ANN model. The procedure is shown in Figure 7.

As shown in Figure 8, the model consists of 5 layers where the input and output layers both contain two units for receiving the dosage vectors and accordingly yielding the SE prediction vectors. The three hidden layers that totally have more than 60 units with enough parameters are used to fit the complex relationships among ontology-based attributes, which are cold and hot here, and the affections thereof.

To train this ANN model, we prepared and washed 150 safe prescriptions from the book *Shanhanzabinglun* and 92 reported prescriptions that frequently cause SE as mentioned before [29–34]. For convenience, we adopted 10-fold cross-validation to train our model, and then we got a convincible result as shown in Section 4.

## 4. Results and Discussion

As seen in Table 1, where bold values highlight the average results obtained in this research, the average accuracy is 87% with a sensitivity rate and a specificity rate of 98% and 17%, respectively. It can be seen in this result that the proposed classifier has a high performance on predicting positive items. Meanwhile, 87% accuracy also proves the high performance of the classifier. However, the low specificity rate means the classifier features poor capability of distinguishing negative items from all data.

This is because the negative data are less than the positive data. The lack of negative data leads to the failure of our ANN model, learning enough knowledge from the provided samples; therefore, the prediction on negative items is more inaccurate. Another reason that should be noted is the features we extracted from medicine data could not represent the typical ones in the classify decision process. Although the other result may be not good as anticipated, the sensitivity rate is out of expectation. The high accuracy on positive items strongly supports our hypothesis.

Hence, the proposed ontology-based SE prediction model is preliminarily verified by the ANN. However, in this procedure, we did not revise the IFs due to the lack of the dataset, resulting in a weak prediction accuracy. Furthermore, the determination of the attributions is relatively broad. In another word, the attributions may be classified into more detailed catalogue such as hot, warm, neutral, cool, and cold. In this way, the ANN could learn more features of the dataset and give more precise predictions. Besides, according to other factors such as different lengths of treatment, it is of great significance that IFs should be evaluated with a weight vector or even tensor, which will influence the results of prescriptions and should be studied in the next stage.

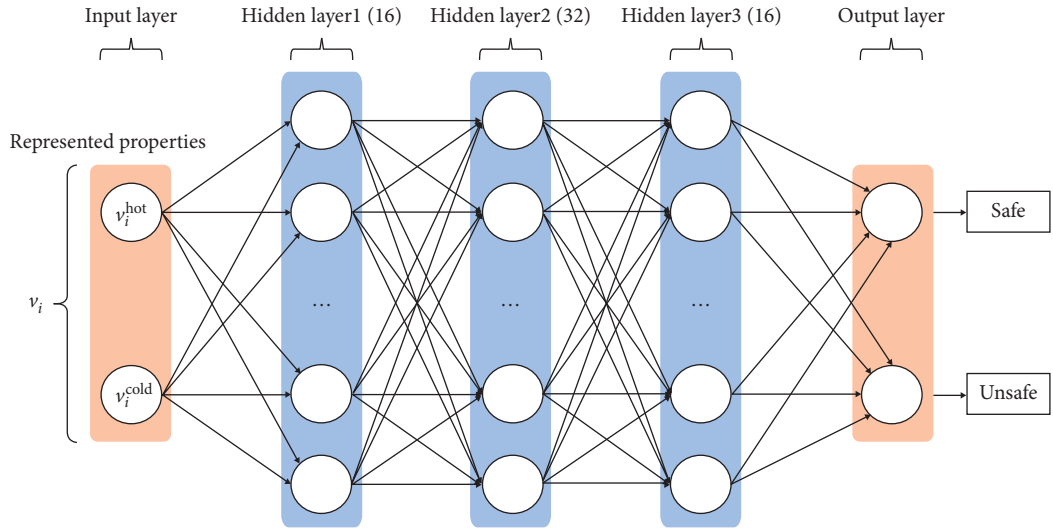

Figure 8: The schematic structure of the ANN and the dataflow.

Table 1: The results of 10-fold cross-validation.

| Fold $N$ | SE | SP | ACC |
|---|---|---|---|
| 1 | 1.00 | 0.00 | 0.92 |
| 2 | 1.00 | 0.33 | 0.92 |
| 3 | 0.92 | 1.00 | 0.92 |
| 4 | 1.00 | 0.33 | 0.92 |
| 5 | 0.91 | 0.00 | 0.88 |
| 6 | 1.00 | 0.00 | 0.79 |
| 7 | 1.00 | 0.00 | 0.92 |
| 8 | 1.00 | 0.00 | 0.88 |
| 9 | 1.00 | 0.00 | 0.79 |
| 10 | 0.95 | 0.00 | 0.79 |
| Average | **0.98** | **0.17** | **0.87** |

## 5. Conclusion

An ontology-based model for AI-assisted medicine side-effect prediction is proposed in this paper. The drug, treatment, and prediction models are established to describe the methodology. In addition, the SE prediction is carried out and verified by the ANN, in which a simplified scheme containing latent attributions (cold and hot) and corresponding indicators (with or without SE) is investigated preliminarily. Clinic data coming from both safe and unsafe prescriptions are adopted to train the ANN and thereafter predict SE. The success of predicting whether a prescription will cause SE demonstrates the simplicity and effectiveness of this work, which should, however, be further improved as a powerful tool to predict more side-effect syndrome.

## Data Availability

The data used to support the findings of this study are available from the corresponding author upon request.

## Conflicts of Interest

The authors declare that they have no conflicts of interest.

## Acknowledgments

This work was supported in part by the National Natural Science Foundation of China under Grant 61370202.

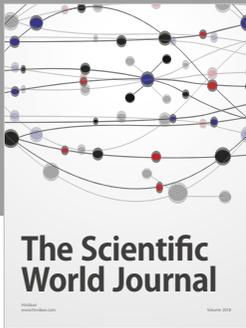
**The Scientific World Journal**

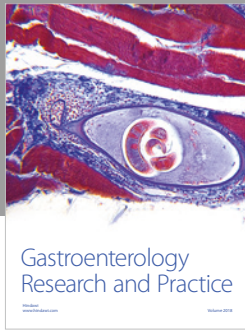
Gastroenterology Research and Practice

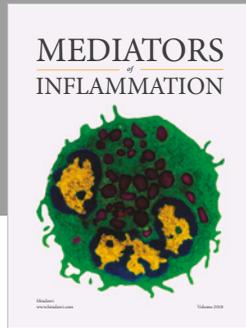
MEDIATORS of INFLAMMATION

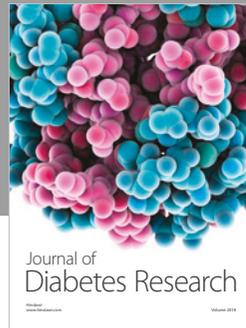
Journal of Diabetes Research

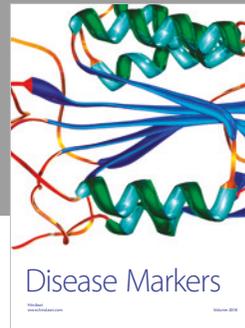
Disease Markers

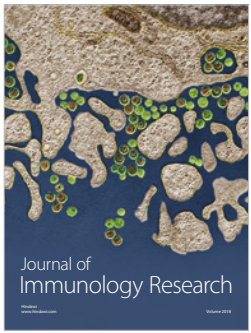
Journal of Immunology Research

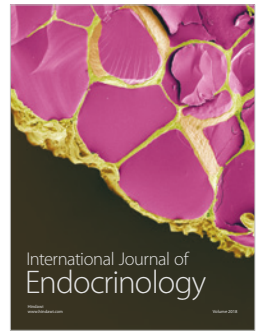
International Journal of Endocrinology

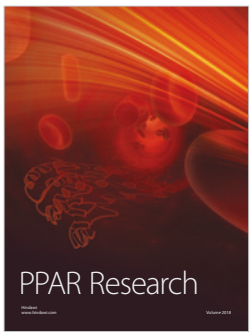
PPAR Research

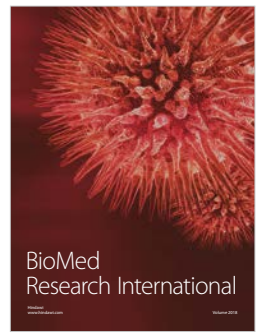
BioMed Research International

Hindawi

Submit your manuscripts at
www.hindawi.com

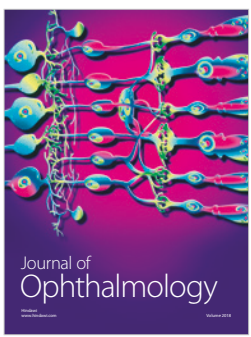
Journal of Ophthalmology

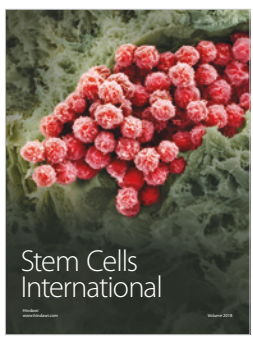
Stem Cells International

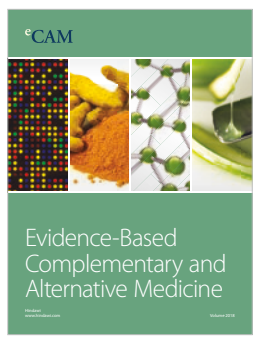
CAM
Evidence-Based Complementary and Alternative Medicine

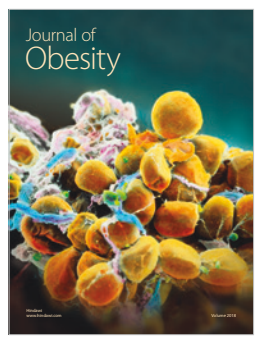
Journal of Obesity

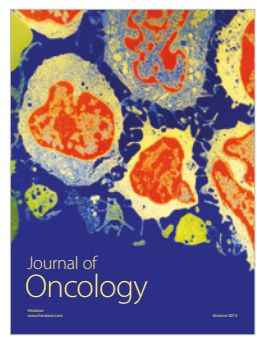
Journal of Oncology

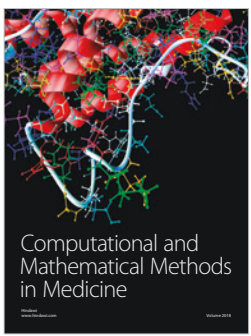
Computational and Mathematical Methods in Medicine

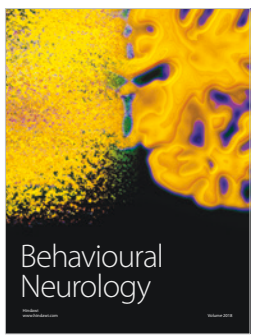
Behavioural Neurology

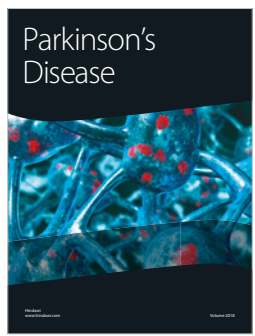
Parkinson's Disease

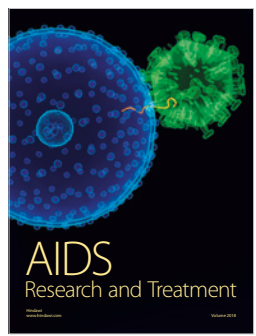
AIDS
Research and Treatment

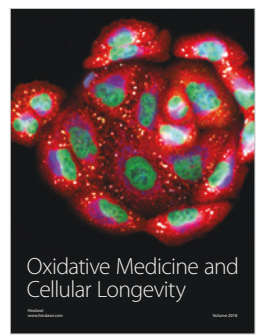
Oxidative Medicine and Cellular Longevity